\documentclass[10pt,twocolumn,letterpaper]{article}

\usepackage[pagenumbers]{cvpr} 
\usepackage[dvipsnames]{xcolor}

\definecolor{cvprblue}{rgb}{0.21,0.49,0.74}
\usepackage[pagebackref,breaklinks,colorlinks,citecolor=cvprblue]{hyperref}
\usepackage{xcolor}
\usepackage{tabularray}
\usepackage{comment}
\usepackage{multirow}

\def\paperID{1097} 
\def\confName{CVPR}
\def\confYear{2024}

\title{Towards Establishing Dense Correspondence on Multiview Coronary Angiography: From Point-to-Point to Curve-to-Curve Query Matching}

\author{Yifan Wu\(^1\)\thanks{Equal contribution.}
\quad Rohit Jena\(^1\)\footnotemark[1]  
\footnotetext{\text{*}Equal contribution.}
\quad Mehmet Gülsün\(^2\)
\quad Vivek Singh\(^2\)
\quad Puneet Sharma\(^2\)
\quad James C. Gee\(^1\)
\\
\(^1\)University of Pennsylvania, Philadelphia, PA, USA\\
{\tt\small \{yfwu, rjena\}@seas.upenn.edu}
{\tt\small \ gee@upenn.edu}\\
\(^2\)Siemens Healthineers, Princeton, NJ, USA\\
{\tt\small \{akif.gulsun, vivek-singh, sharma.puneet\}@siemens-healthineers.com\quad}
}

\begin{document}
\maketitle
\begin{abstract}
Coronary angiography is the gold standard imaging technique for studying and diagnosing coronary artery disease. However, the resulting 2D X-ray projections lose 3D information and exhibit visual ambiguities. In this work, we aim to establish dense correspondence in multi-view angiography, serving as a fundamental basis for various clinical applications and downstream tasks. To overcome the challenge of unavailable annotated data, we designed a data simulation pipeline using 3D Coronary Computed Tomography Angiography (CCTA). We formulated the problem of dense correspondence estimation as a query matching task over all points of interest in the given views. We established point-to-point query matching and advanced it to curve-to-curve correspondence, significantly reducing errors by minimizing ambiguity and improving topological awareness. The method was evaluated on a set of 1260 image pairs from different views across 8 clinically relevant angulation groups, demonstrating compelling results and indicating the feasibility of establishing dense correspondence in multi-view angiography.
\end{abstract}  
\section{Introduction}
\label{sec:intro}

Cardiovascular disease, particularly coronary artery disease, is a worldwide concern that imposes a substantial burden on patients and healthcare systems \cite{smith2013moving}. In the United States alone, more than one million ICA procedures are conducted each year \cite{virani2021heart}. The gold standard for coronary artery disease imaging is invasive coronary angiography (ICA) \cite{ryan2002coronary}, which provides valuable information for assessing coronary arteries to facilitate diagnoses and intervention planning. 

Angiography is a form of X-ray that uses fluoroscopy after contrast injection to identify narrowed or blocked arteries. However, since 2D X-ray projection does not preserve 3D information, the resulting images often exhibit visual ambiguities, such as foreshortening, overlapping branches, and eccentric stenosis. Therefore, angiographies are typically acquired from multiple angulations in order to better assess coronary anatomy and stenoses \cite{green2004three,gollapudi2007utility}.

In this work, our aim is to establish dense correspondence in multi-view angiography. This will serve as a fundamental basis for various clinical applications and the development of downstream algorithms. For example, these applications include constructing image-based Fractional Flow Reserves (FFR), which involves generating a 3D anatomical model of the coronary from angiographies \cite{morris2020angiography}. The dense correspondence framework can serve as a foundational model of the 2D X-ray images, providing an understanding of the 3D shape priors, which can further be utilized for the development of downstream algorithms.


\begin{figure*}[t]
\centering
    \includegraphics[width=\textwidth]{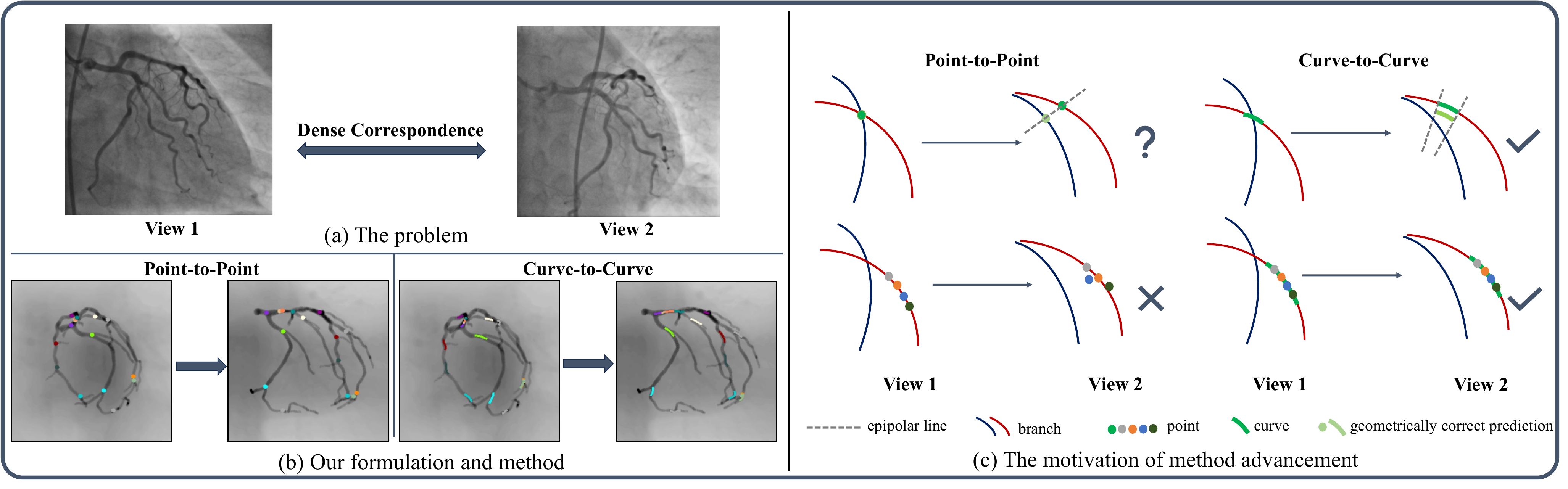}
\caption{\textbf{Schematics.} Panel (a) illustrates the problem we aim to solve—establishing dense correspondence between images from different angles. Panel (b) presents our formulation of the problem and our methods: we define the correspondence problem as a query matching task, solving both point-to-point and curve-to-curve correspondences. Panel (c) demonstrates the motivation for advancing from point-to-point to curve-to-curve queries. Leveraging branch context allows for reduced ambiguity from branch overlapping (first row). Also, treating the curve as a query, rather than as an unordered set of points, reduces deviation from the actual branch path (second row).}
\label{fig:schematic}
\end{figure*}

Previous studies have focused on solving sparse correspondence in angiography, such as tracking the catheter tip, stenoses, or vessel segments \cite{zhang2019cascade,huang2022robust,zhang2017vessel}. However, research on developing dense correspondence is currently limited. The main challenge impeding the development of such techniques is the absence of annotated data, as labeling each point of the target structure across different views is labor-intensive. To tackle this challenge, we designed a pipeline to simulate 2D data from 3D Coronary Computed Tomography Angiography (CCTA). To achieve this, we rendered clinically relevant 2D projection images using the digitally rendered radiographs (DRR) technique from 3D data. By utilizing the labels available on 3D data, such as the entire coronary centerline points, we projected these labels onto the 2D images, which allowed us to acquire dense correspondence ground truth between different views.

The task of vision correspondence can be divided into two streams, namely sparse and dense, with each stream being designed to solve different applications. Sparse methods typically first detect a pool of salient points \cite{detone2018superpoint} and then perform feature matching to obtain correspondence \cite{sarlin2020superglue}. On the other hand, dense correspondence predicts a displacement field, which usually models small changes and relies on local smoothness, such as optical flow \cite{teed2020raft} or registrations in the context of medical images analysis \cite{wu2022nodeo}. However, neither approach is well-suited for projected angiographic images due to the large camera pose differences and the need for correspondence at any probed points. Inspired by COTR \cite{jiang2021cotr}, which unified sparse and dense correspondence within a single framework, we formulated the dense correspondence problem as a query task. We built a point-to-point query matching framework, which predicts the corresponding points on the target image given the reference image, query points, and the target image. In this way, our framework is capable of achieving correspondence over any point of interest between two views. 

Our problem presents several domain-specific challenges. Unlike natural images, angiograms lack rich, discriminative background textures and image features, making correspondence prediction more difficult, as shown in Fig.~\ref{fig:schematic} (a).
Furthermore, as shown in Fig.~\ref{fig:schematic} (c), the branched structure of coronary arteries introduces ambiguities in 2D projections. For instance, when multiple arteries overlap at a given pixel, it becomes unclear which artery is being queried. It is also challenging to determine whether a junction point in 2D is a real bifurcation or branch occlusion in 3D. This ambiguity cannot be resolved with geometric priors (i.e., epipolar constraints) since both arteries intersect with the epipolar line by construction. Additionally, treating neighboring points independently with point queries lacks topological awareness of their continuous order. These challenges have motivated us to leverage more branch context near the probing point. Consequently, we have extended our framework from point-to-point query matching to curve-to-curve query matching. During training, we developed models for both point and curve correspondence. During inference, users can query points and, optionally, draw a small segment to provide additional context. The curve prediction will then refine the point matching for enhanced accuracy.

Our method is evaluated quantitatively and qualitatively on 8 different clinically relevant angulation groups \cite{di2005coronary}. We also conducted experiments on centerline tracing to demonstrate that tracing accuracy can be improved by incorporating correspondences from a second view. Additionally, we qualitatively evaluated this approach on real angiograms.  In summary, our contributions in this work are: 
\begin{enumerate}
    \item We are the first to attempt solving the dense correspondence problem in multiview angiographies, designing a data simulation pipeline to enable training, and formulating the correspondence as a query problem.
    \item We advanced our framework from point-to-point correspondence to curve-to-curve correspondence, reducing ambiguity and improving topological awareness,  which results in a 25\% overall reduction in prediction errors.
    \item Our results demonstrate the feasibility of establishing dense correspondence from 2D multi-view images both quantitatively and qualitatively, paving the way for future works such as fully automatic 3D reconstruction.
\end{enumerate}

\section{Related Work}
\label{sec:relatedwork}
\subsection{Vessel correspondence and reconstruction}
Previous studies on coronary angiograms have focused on solving sparse temporal correspondence in \textit{single-view} angiograms, such as tracking catheter tip and stenoses, or segments of vessels~\cite{zhang2017vessel,zhang2019cascade,huang2022robust}.
Several studies on establishing dense multi-view correspondences are motivated by forward-projecting an estimated 3D centerline onto 2D views~\cite{cong2013energy,cong2015quantitative,yang2014external}.
These methods aims to update the estimated 3D centerline by minimizing an energy function of its 2D projections. 
Establishing 2D correspondence from 3D then becomes straightforward.
However, these approaches require manually initialized 3D centerlines to avoid local optima. 
Other methods aim to establish 2D correspondences and backproject these correspondences to 3D.
Kunio \etal~\cite{kunio2018vessel} propose a method for reconstructing vessel centerline in 3D from a pair of non-orthogonal angiographic images. 
The method enhances the contrast of the images, followed by manually thresholding the image to show only the stented artery and the side branch.
For non-isocentric images, 2D correspondences are first established manually.
This method involves a human-in-the-loop approach, and a relatively simplified setup for correspondence finding with a single vessel segment.
Kalmykova \etal~\cite{kalmykova2018approach} propose a point-to-point correspondence approach by considering epipolar constraints.
However, because the epipolar line intersects the coronary arteries at multiple centerline locations, the user manually intervenes and selects the point that best describes the correspondence.
In contrast to these methods, we consider a more realistic and significantly harder problem setup, where the entire coronary artery system is considered, the images are not isocentric.
Moreover, epipolar constraints are not sufficient to resolve correspondence in our case due to different arteries occluding each other in the image plane (Fig.~\ref{fig:schematic}).
Owing to the fully automatic approach and 
aforementioned 
requirements,~\cite{kunio2018vessel,kalmykova2018approach} are not applicable in our setting.

\subsection{Representations of correspondence}
The task of vision correspondence can be divided into two streams, namely sparse and dense, with each stream being designed to solve different applications.
Sparse correspondence typically attempts to identify a set of keypoints that correspond to matching features or structures in the image pair~\cite{lowe1999object,bay2008speeded,rublee2011orb,sarlin2020superglue,luo2019contextdesc}. 
This is used to minimize some alignment metric, like relative camera motion.
In contrast, the dense correspondence setup attempts to map every pixel in a query image to some pixel in the target image~\cite{liu2010sift,sun2014quantitative,zhou2017unsupervised}, which usually models small changes and relies on local smoothness.
This formulation is commonly used in dense tracking~\cite{lucas1981iterative,patrick2021keeping,rocco2018neighbourhood}, optical flow ~\cite{teed2020raft,zhu2020deformable,xu2022gmflow}, and image registration problems~\cite{chen2022transmorph,mok2020large,gee1993elastically,gee1998elastic,balakrishnan2019voxelmorph,yoo2017ssemnet}.
However, neither of these two streams is directly suitable to our problem. Since the vascular centerlines are distributed densely in angiograms, we cannot utilize the sparse correspondence setup.
However, most pixels (\eg background points) do not have meaningful correspondences, and the angulation differences in the images is large, so the dense correspondence setup does not apply as well.

This calls for representing the correspondence problem as a query matching paradigm, where a dense set of queries can be mapped to a set of targets.
Following this query-target paradigm for dense correspondence, COTR~\cite{jiang2021cotr} propose a framework for specifying query points in a query image and predicting corresponding points in the target image.
However, establishing point correspondences can be an underconstrained problem, especially in situations with a lack of distinct image features or where enforcing topological relationships between correspondences is necessary.
This avenue is explored by extending the correspondence problem to other primitives such as lines~\cite{faugeras1995geometry,1211499,pautrat2023gluestick,pautrat2021sold2,zhao2022superline3d}, aiming to enhance the reliability of the descriptor.

In our problem, coronary arteries have curvature to accomodate the shape of the heart, and can be tortuous which can affect clinical procedures like angiography.
This behavior necessitates the use of curve representation in a query matching paradigm, which we explore in this work. 

\section{Method}
\label{sec:method}

\subsection{Data Simulation}

\begin{figure}[t]
\centering
    \includegraphics[width=0.48\textwidth]{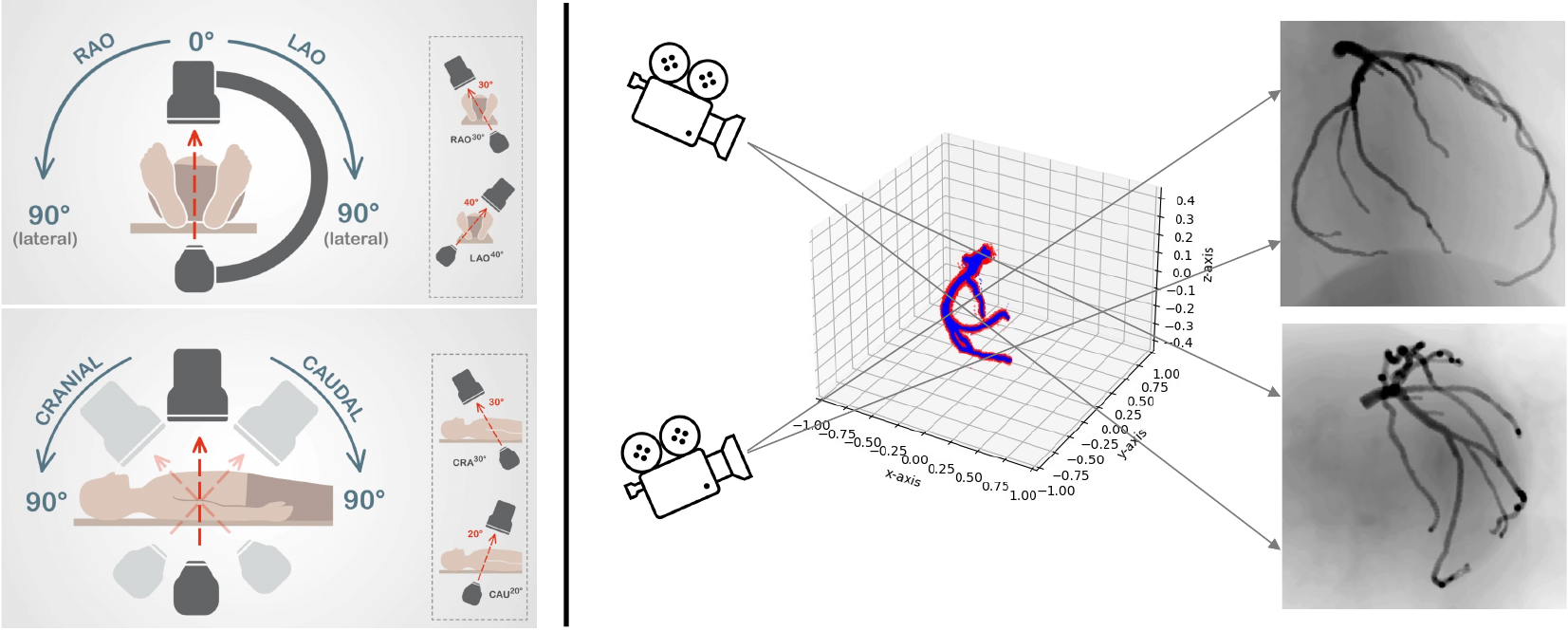}
\caption{\textbf{Data simulation workflow.} The left figure of the C-arm is adapted from medmastery.com.}
\label{fig: data_simulation}
\end{figure}

One of the primary obstacles encountered in the development of dense correspondence lies in the unavailability of annotated data. 
Our approach involves tackling this challenge by simulating 2D X-ray-like images from CCTA. Specifically, we produced 2D images from clinically relevant projections by rendering the aforementioned 3D CT. With the 3D coronary centerlines and bifurcations, and stenosis markers at our disposal, we were able to project the 3D labels onto 2D images using the relevant camera model.

In the clinical workflow where the patient is positioned at the center, two axes of rotation of the angiogram are involved. The first rotation is along the left anterior oblique-right anterior oblique (LAO-RAO) axis, which we denote as $\alpha$. The second rotation is along the cranial-caudal 
axis, which we denote as $\beta$. The projection is subsequently defined by the normal $\vec{n}$ of the detector plane. 

\begin{equation}
\vec{n} = [\sin (\alpha) \cos (\beta),-\cos (\alpha) \cos (\beta), -\cos (\alpha) \sin (\beta)]
\end{equation}

\label{method:data simulation}

We produced images using 8 projection groups according to the clinical guideline specified in~\cite{di2005coronary}. We rotate $\alpha$ and $\beta$ in increments of 5°, resulting in the number of images for each group being: (A) LAO 40-50°, caudal 25-40°, 3 images; (B) RAO 5-15°, caudal 30°, 3 images; (C) RAO 30-45°, caudal 30-40°, 12 images; (D) RAO 5-10°, cranial 35-40°, 6 images; (E) LAO 30-45°, cranial 25-35°, 12 images; (F) Lateral ±, caudal-cranial 10-30°, 10 images; (G) LAO 45-60°, 4 images; (H) RAO 30-45°, 4 images. We rendered the left coronary artery (LCA) and the right coronary artery (RCA) separately, resulting in $63 \times 2$ images for one subject. Our final rendered images are $512 \times 512$ pixels with a resolution of $0.58$ millimeter per pixel. 
An illustration of our data simulation pipeline is shown in Fig.~\ref{fig: data_simulation}.

After rendering, we can get 2D label point $X_p$ by projecting 3D label point $X_w$ to 2D images using the projection model \cite{Hartley2004}:
\begin{equation}
X_p = K \left[\begin{array}{c|c}
R & t
\end{array}\right] X_w
\end{equation}
where $X_w$ is the 3D label point in world reference system, $R$ and $t$ are known as the extrinsic parameters, and $K$ is camera matrix, i.e., the intrinsic parameters.


\begin{figure*}[!t]
\centering
\includegraphics[width=0.98\textwidth]{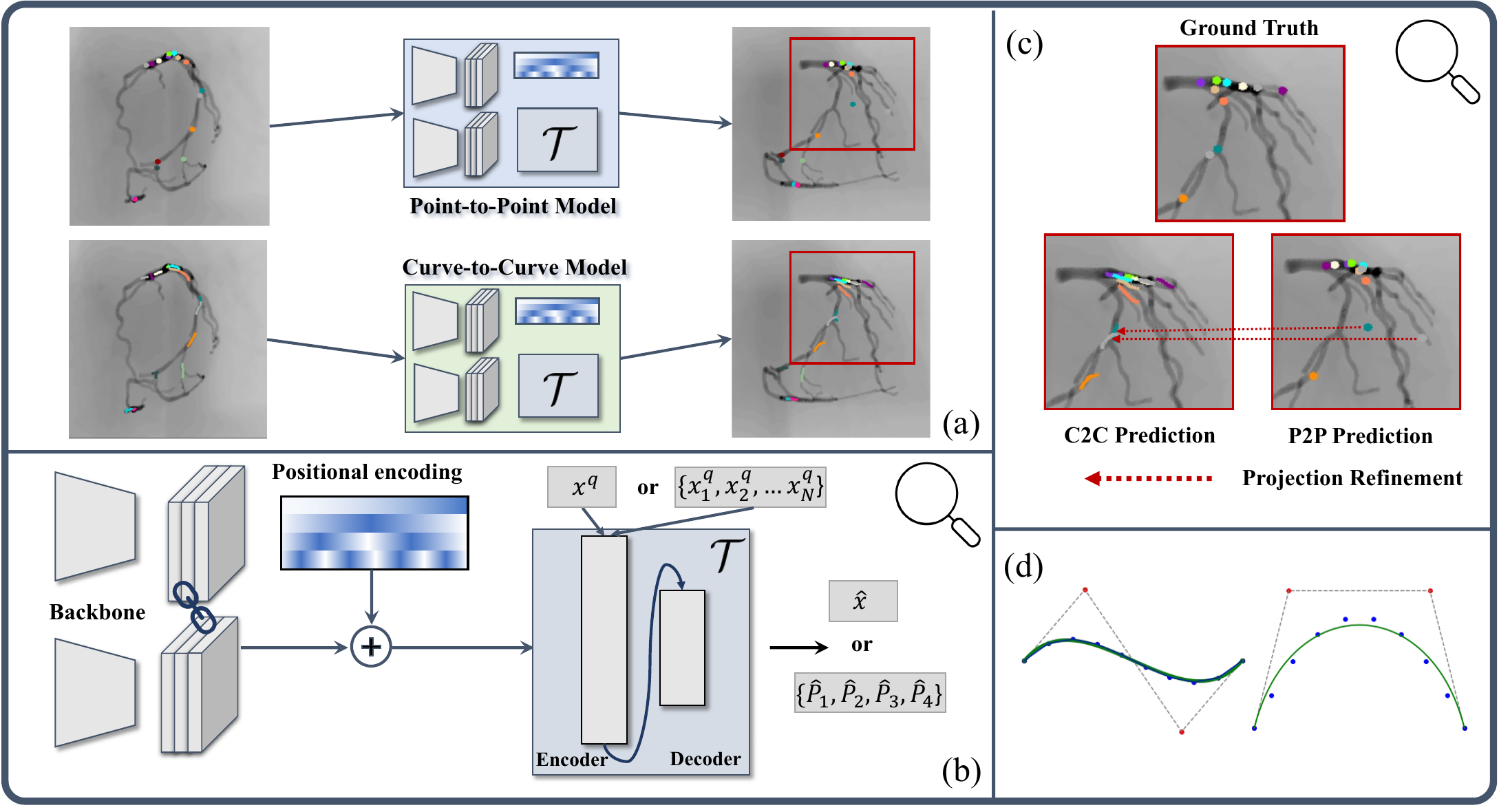}
\caption{\textbf{An overview of our method.} During training, we trained a Point-to-Point (P2P) model and a Curve-to-Curve (C2C) model, as shown in Panel (a). Panel (b) demonstrates an enlarged illustration of the correspondence transformer. The framework begins with a backbone to process each image and generate feature maps. These feature maps are then concatenated and merged with positional encodings, which are fed into a transformer along with the query. Panel (c) shows an enlarged view of the red regions in Panel (a). During inference, one can use P2P, C2C, or project point to curve for refinement. Panel (d) illustrates our curve parameterization.}
\label{framework}
\end{figure*}

\subsection{Coronary Correspondence Transformer}
\label{method:point-to-point}

Our goal is to learn a transformer network $F$ that learns a mapping, $F: (x^{q} \mid I_{ref}, I_{tgt}) \rightarrow \hat{x}$, to predict the correspondence point $\hat{x}$ on the target image $I_{tgt}$, given the query coordinate $x^{q}$ on the reference image $I_{ref}$.

An overview of our framework is shown in Fig.~\ref{framework}. First, we use ResNet-50 \cite{he2016deep} to obtain feature representations of input images $I_{ref}$ and $I_{tgt}$, resulting in feature maps $z_{ref}$ and $z_{tgt}$. These feature maps are then fed into the transformer network with positional encoding. 
Specifically, $z_{ref}, z_{tgt}$ $\in \mathbb{R}^{C \times H \times W}$, here $H=W=16, C=256$. Then we concatenate two feature maps side by side, forming a feature map $z$ in size of $256 \times 16 \times 32$. This concatenation is crucial because it enables the transformer encoder to establish relationships between locations within and across images through self-attention and cross-attention mechanisms. 
The next step is to embed the positions of the feature map. We use linear positional encoding as suggested in \cite{jiang2021cotr}, denoted by $P$, where $P$ utilizes a linear increase in frequency. For a given point $x$, 
\begin{equation}
\small
P(x)=\left[p_1(x), p_2(x), \ldots, p_{\frac{C }{4}}(x)\right]
\end{equation}
where $p(x)=\left[\sin \left(k \pi x^{\top}\right), \cos \left(k \pi x^{\top}\right)\right]$, $C=256$ is the channel number of $z$.
After obtaining the context feature map $z$ and positional encoding of the domain of image $I_{ref}$, we pass it through a transformer encoder.
The transformer decoder interprets the results along with the positional embedding of the query coordinates. The output of the transformer decoder is then processed by a fully connected layer to estimate the corresponding coordinates.

\begin{equation}
\small
\hat{x}=F_\theta \left((x^{q} \mid I_{ref}, I_{tgt})\right)=FC\left(T_D\left(P(x^q), T_E(z \cdot P (\Omega))\right)\right)
\end{equation}
where $T_E$ and $T_D$ denote transformer encoder and decoder, $FC$ denotes fully connected layer, $\Omega$ is the domain of image $I_{ref}$, $|\cdot|$ denotes concatenation on the spatial dimension. Here both the $T_E$ and $T_D$ contain six layers. Each encoder layer includes an eight-head self-attention module, whereas each decoder layer includes an eight-head encoder-decoder attention module. We use a 3-layer multilayer perceptron for $FC$ to predict $\hat{x}$ from the 256-dimensional latent vector.

For learning objectives, we penalize the euclidean distances of  predicted coordinates and ground-truth coordinates. We also use the cycle loss to enforce the correspondences to be cycle-consistent. The loss functions are as follows: 

\begin{equation}
\small
\begin{aligned}
\mathcal{L}_{corr} = \sum_{i=1}^N\left\|x^{t}_{i}-F_{\theta}\left(x^{q}_{i} \mid I_{ref}, 
 I_{tgt}\right)\right\|_2^2 
 \end{aligned}
 \label{eq:l_corr}
\end{equation}

\begin{equation}
\small
\begin{aligned}
\mathcal{L}_{cycle} = \sum_{i=1}^N \|x^{q}_{i}- F_{\theta} (F_{\theta} (x^{q}_{i} \mid I_{ref}, 
 I_{tgt})  \mid I_{ref}, I_{tgt})\|_2^2
 \end{aligned}
\end{equation}

where $N$ is the number of query points, $x^{q}_{i}$ is the query position and $x^{t}_{i}$ is the ground-truth of target position. The final objective is $\arg \min _\theta \left(\mathcal{L}_{corr}+\lambda \mathcal{L}_{cycle}\right)$.
The $\lambda$ is a hyper-parameter to weight $\mathcal{L}_{corr}$ and $\mathcal{L}_{cycle}$. The backbone ResNet-50 is pre-trained on Imagenet. The learning rates are 1e-5 for the  backbone and 1e-4 for the remaining transformer network. During training, we randomly sample $N=100$ points for each data sample. The image input is of size $320$x$320$ pixels. 


\subsection{Extension to Curve to Curve Correspondence}
The point to point framework has two limitations.
First, the framework treats sequential points on a segment independently.
There is no topological awareness either in the model or in the MSE loss function (Eq.~\ref{eq:l_corr}).
This leads to predictions that produce tiny MSE errors but violate the topological structure of the vessel and causing drift (Fig.~\ref{fig:schematic}(c)).
Second, point queries lead to ambiguity in the target prediction if the query lies on the intersection of two vascular segments in the 2D image.
To leverage this additional topological structure in the domain and to prevent ambiguities in prediction for overlapping vessels, we extend the formulation by allowing a \textit{query segment} instead of a query point.

A curve segment is represented by $N$ ordered points $\mathcal{C}_q = \{\mathbf{x}^q_1, \mathbf{x}^q_2, \ldots \mathbf{x}^q_N\}$, where $N$ is chosen as a hyperparameter.
Our goal now becomes to learn a transformer network $F_\mathcal{C}$ that learns the mapping $F_\mathcal{C}: (\mathcal{C}_q | I_{ref}, I_{tgt}) \rightarrow \hat{\mathcal{C}}$.
For consistency, we use the same architecture as the point-to-point correspondence.
However, to incorporate multiple ordered inputs, we perform the extra preprocessing.
First, we compute the positional encoding of each query point to create an ordered list of encoded vectors $\{ P(\mathbf{x}^q_1), P(\mathbf{x}^q_2), \ldots P(\mathbf{x}^q_N) \}$.
Next, we concatenate these vectors into a single embedding and pass it through a 3-layer MLP to produce a `segment embedding'. 
This segment embedding is then used to produce a target embedding which is decoded into a target segment.
However, if the target segment is predicted as a set of points, it may violate the ordering of the segment if trained with the MSE loss.
To enforce the topology of a curve, we predict the control points of a Bezier curve instead. 
This enforces the segment topology by construction, and allows us to train the control points to align with the target points.
We describe the training of the Curve to Curve below.

Consider a query segment $\mathcal{C}_q$ and its corresponding target segment $\mathcal{C}_t$.
We fit query and target Bezier curves $\mathcal{B}_q, \mathcal{B}_t: [0, 1] \rightarrow \mathbb{R}^2$ from the segments using the Global Curve Interpolation ~\cite{piegl1996nurbs} algorithm.
Given a target curve prediction $\hat{\mathcal{B}}$, we define the curve-to-curve as the Chamfer distance between points on the predicted Bezier curve and the target segment.
\begin{equation}
    \mathcal{L}_{c2c}(\hat{\mathcal{B}}, \mathcal{C}_t) = \sum_{\mathbf{x} \in \mathcal{C}_t} \min_{u \in [0, 1]} d(\hat{\mathcal{B}}(u), \mathbf{x}) + \int_{0}^{1} \min_{\mathbf{x} \in \mathcal{C}_t} 
 d(\hat{\mathcal{B}}(u), \mathbf{x}) du
\end{equation}
where $d(., .)$ is the squared L2 distance between two points.
We also employ a supervised loss between the predicted control points and the target control points
\begin{equation}
    \mathcal{L}_{sup}(\hat{\mathcal{B}}, \mathcal{B}_t) = \sum_{i=1}^N \sum_{i=1}^{4} d(\mathbf{P}^i_{\mathcal{B}^t}, \mathbf{P}^i_{\hat{\mathcal{B}}})
\end{equation}

\begin{figure*}[!t]
\centering
\includegraphics[width=\textwidth]{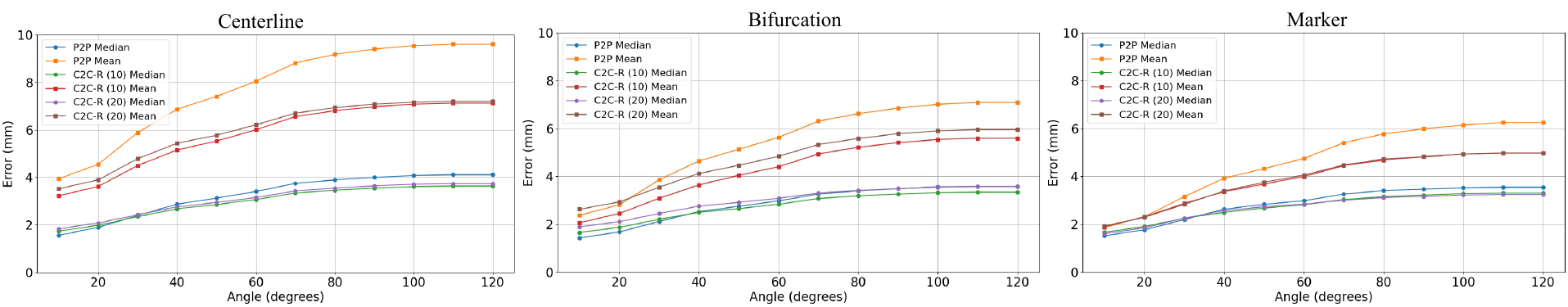}
\caption{\textbf{Mean and Median Error (mm) of All Centerline Points within Certain Angle Differences Between Reference and Target Images.} A Comparison of Point-to-Point, Curve-to-Curve (10 Points) Refined, and Curve-to-Curve (20 Points) Refined Results.}
\label{fig:angle_error}
\end{figure*}

To incorporate the cycle loss, we sample ordered points from $\hat{\mathcal{B}}$ as the input to the network, and predict a new Bezier curve denoted by $\hat{\mathcal{B}}_{cyc}$, i.e. $\hat{\mathcal{B}}_{cyc} = F_{\mathcal{C}}( S(F_{\mathcal{C}}( \mathcal{C}_q | I)) | I)$, where $S$ is a sampling operation to convert the Bezier control points into an ordered list of points on the segment.
The Cycle consistency loss is then given by
\begin{equation}
    \mathcal{L}_{cyc} = \mathcal{L}_{c2c}(\hat{\mathcal{B}}_{cyc}, \mathcal{C}_q) + \lambda_s \mathcal{L}_{sup}(\hat{\mathcal{B}}_{cyc}, \mathcal{B}_q)
\end{equation}
where $\lambda_s$ is a hyperparameter to weigh the relative importance of $\mathcal{L}_{sup}$ and $\mathcal{L}_{c2c}$.
During training the curve-to-curve network, we randomly sample $N=100$ curves from each data sample. The image feature embedding network and other training configurations are the same as the point-to-point network introduced in Sec~\ref{method:point-to-point}.

\begin{figure}[t]
\centering
\includegraphics[width=0.45\textwidth]{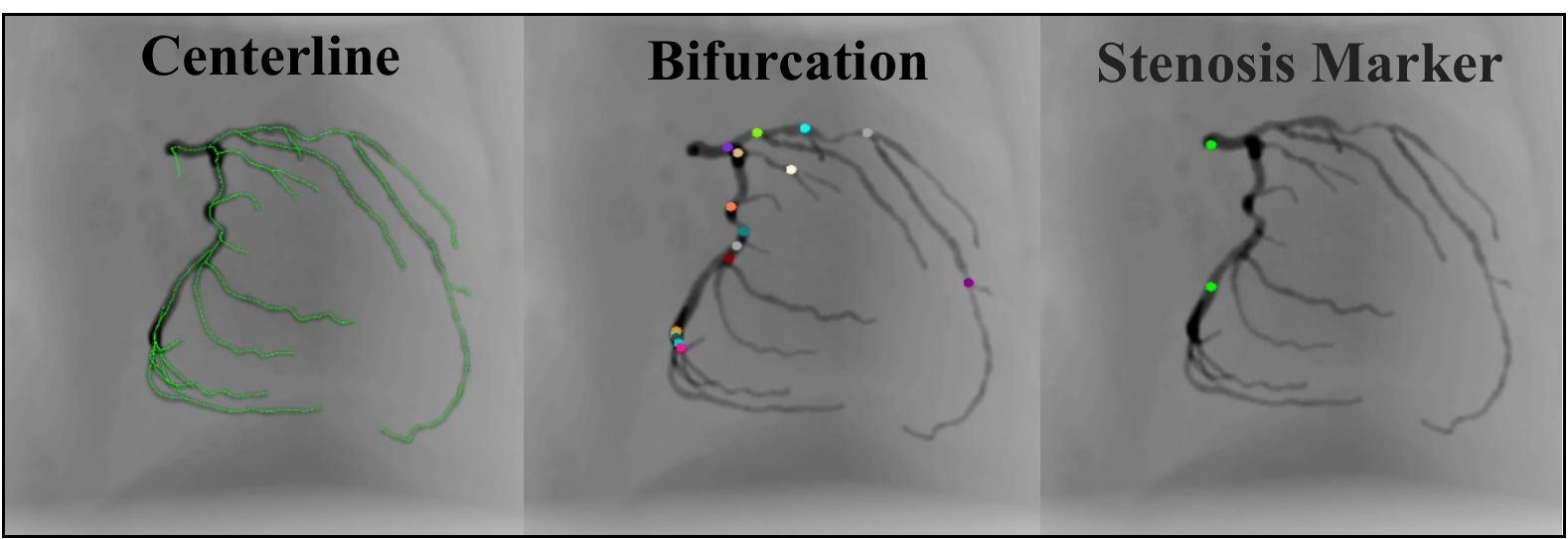}
\caption{\textbf{An example of simulated data and labels}}
\label{fig: data-vis}
\end{figure}

\subsection{Inference}
For the Point-to-Point (P2P) model, inference is straightforward - a query point is fed into the model which predicts its correspondence point.
We denote this prediction as the Point-to-Point prediction.
For the Curve-to-Curve (C2C) model, we provide an ordered sequence of points, which we call a \textit{waypoint}.
This query waypoint is fed into the C2C model which outputs the parameters of a Bezier curve, representing the target waypoint.
The C2C model can be used in two ways. 
First, the curve prediction can be interpreted as a new task in and of itself.
However, this does not enable us to directly compare it with the P2P model.
Therefore, we propose a second way which is to use the predicted point from the P2P model and project it to the nearest point on the curve predicted by the C2C model.
We denote this projected point as the \textbf{C2C refined (C2C-R)} prediction.
The fidelity of the C2C model can now be measured indirectly by its ability to localize the P2P prediction correctly.

\begin{figure*}[t]
\centering
\includegraphics[width=0.95\textwidth]{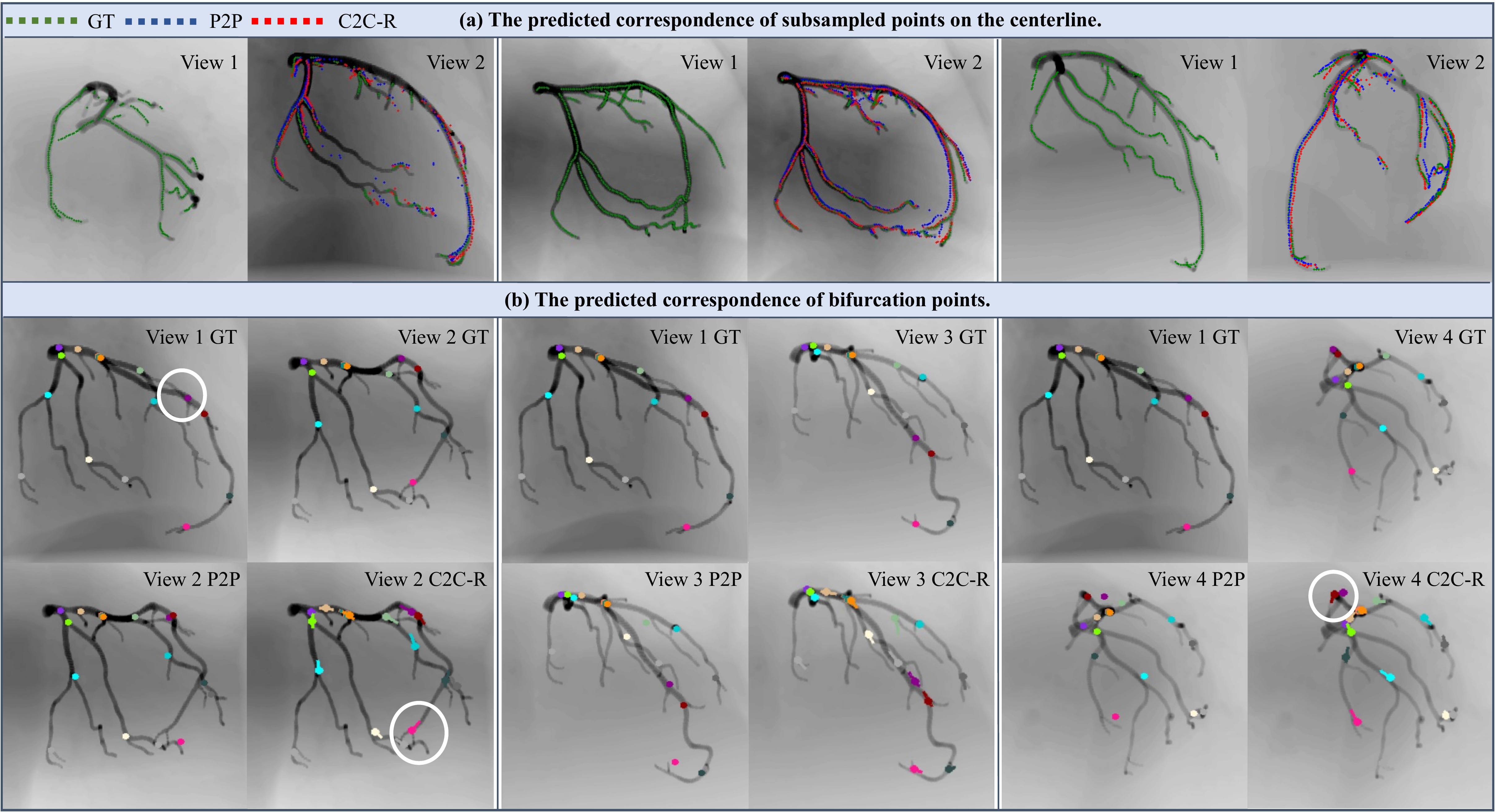}
\caption{\textbf{The qualitative result of correspondence between different views.} Panel (a) displays the correspondence of centerline points. In each example pair of images, the left one shows the reference image with ground-truth labels in \textcolor{green}{green}. The right one illustrates the predicted correspondence: from the Point-to-Point (P2P) model in \textcolor{blue}{blue}, and the Curve-to-Curve refined (C2C-R) results in \textcolor{red}{red}. Panel (b) demonstrates the correspondence of bifurcation points between View 1 and View 2, View 1 and View 3, and View 1 and View 4, organized into three groups of columns. The first row displays reference and target images with the ground truth point. The second row shows the predicted correspondence: the P2P model predictions (P2P) on the left, and the C2C-R refined results on the right. We highlight some regions of interest with white circles. In View 1 GT, the purple point, lying at the junction of two branches, serves as an example of a source of ambiguity. The highlighted areas in View 2 C2C-R and View 4 C2C-R demonstrate the improvement compared to P2P results.
}
\label{fig: eval-demo}
\end{figure*}

\section{Experiments}
\label{sec:experiment}

\subsection{Data}
In this study, we utilized 99 CCTA volumes, all from different patients. We randomly selected and reserved 5 volumes for validation, 10 for testing, and employed the remaining 84 volumes for training. 
As described in Sec.~\ref{method:data simulation}, we projected each volume to 126 (=$63 \times 2$) images. We utilized all centerline points of the CCTA, including the topology of branches that define the bifurcation points. Additionally, in most cases, stenosis markers were annotated, with start and end points of the stenosis. Finally, there are $1260$ images in the testing set, $630$ each of LCA and RCA. In the testing set, LCA has an average of 1603 centerline points and 16 bifurcation points per image, with 8 subjects having 2 stenosis markers. RCA has an average of 1166 centerline points and 10 bifurcation points per image, with 4 subjects having 2 stenosis markers. An example is shown in Fig.~\ref{fig: data-vis}.

\begin{table}
\centering
\small
\begin{tblr}{
  width = 0.5\textwidth,
  colspec = {X[c]X[c]X[c]X[c]X[c]X[c]X[c]},
  cell{2}{1} = {c=7}{},
  cell{3}{1} = {r=2}{},
  cell{5}{1} = {r=2}{},
  cell{7}{1} = {r=2}{},
  cell{9}{1} = {c=7}{},
  cell{10}{1} = {r=2}{},
  cell{12}{1} = {r=2}{},
  cell{14}{1} = {r=2}{},
}
\hline
\hline
Query & Method & $\leq 10^{\circ}$ &$\leq 30^{\circ}$ &$\leq 50^{\circ}$ &$\leq 70^{\circ}$ &$\leq 90^{\circ}$ \\
\hline
\textbf{Mean}            &   &        &        &       &       &        \\
\hline
Centerline      & P & 3.95   & 5.89   & 6.87  & 8.05  & 9.18   \\
                & C & \textbf{3.23}   & \textbf{4.50}   & \textbf{5.53}  & \textbf{6.56}  & \textbf{6.98}  \\
\hline
Bifurcation     & P & 2.37   & 3.86   & 5.13  & 6.30  & 6.85   \\
                & C & \textbf{2.06}   & \textbf{3.09}   & \textbf{4.04}  & \textbf{4.94}  & \textbf{5.41}   \\
\hline
Stenosis Marker & P & \textbf{1.87}   & 3.16   & 4.32  & 5.41  & 5.99   \\
                & C & 1.90   & \textbf{2.87}   & \textbf{3.68}  & \textbf{4.45}  & \textbf{4.82}   \\
\hline
\textbf{Median}         &   &        &        &       &       &        \\
\hline
Centerline      & P & \textbf{1.56}   & 2.40   & 3.13  & 3.75  & 4.00   \\

                & C & 1.74   & \textbf{2.35}   & \textbf{2.86}  & \textbf{3.34}  & \textbf{3.55}   \\
\hline
Bifurcation     & P & \textbf{1.43}   & \textbf{2.12}   & 2.75  & 3.26  & 3.49   \\
                & C & 1.66   & 2.21   & \textbf{2.66}  & \textbf{3.08}  & \textbf{3.26}   \\
\hline
Stenosis Marker & P & \textbf{1.52}   & \textbf{2.19}   & 2.83  & 3.26  & 3.47   \\
                & C & 1.67   & 2.25   & \textbf{2.68}  & \textbf{3.03}  & \textbf{3.22}   \\
\hline
\hline
\end{tblr}
\caption{\textbf{The mean and median error (mm) of all centerline points within certain angle differences between reference and target images.} ``P'' refers to Point-to-Point; ``C'' refers to Curve-to-Curve (10 points) Refined Prediction.}
\label{table:angle_error}
\end{table}

\subsection{Quantitative and Qualitative Analysis}
In our experimental setup, each image in the testing set served as the reference image and was randomly paired with another image from the target angulation group to act as the target image. Testing was conducted separately on all stenosis markers, bifurcations, and centerline points. The mean and median errors for all queried points within certain angles are reported in Table~\ref{table:angle_error} and Fig.\ref{fig:angle_error}. 
We observed that C2C-R predictions are consistently better across all centerline points, bifurcation points, and stenosis markers. This improvement is particularly significant when the angle difference is large, highlighting a more challenging correspondence problem where the context of the query plays a more important role. Overall, C2C-R reduces errors by 25\% (from 9.61 to 7.13 as in Table~\ref{tab:ablation}).
Furthermore, while the majority of points (approximately 75\%) exhibited low errors, a minority (approximately 25\%) caused substantial errors, leading to higher mean errors compared to the median distances. Notably, centerline points presented larger distance errors compared to stenosis markers and bifurcation points, especially those located near the ends of branches. Qualitative examples of data and predictions are shown in Fig.\ref{fig: eval-demo}. Here, it is evident that the P2P model yields reasonable predictions, even in cases of large angle differences. Additionally, the C2C model further refines these predictions. 
The quantitative and qualitative analysis demonstrates our framework's ability to comprehend coronary structures and validates the feasibility of establishing correspondences in multi-view angiographic images.

\subsection{Ablation Study on Waypoint Size}
One of the parameters introduced in the curve-to-curve paradigm is the number of ordered points provided as the query segment.
More number of points provide more context for the curve correspondence task, but providing too many points can lead to curves that deviate from the Bezier curve assumption, making modelling difficult.
Moreover, in a real clinical setting, the clinician would prefer to provide as few points as possible.
With these considerations, we consider and compare two waypoint sizes - 10 and 20 points. 
We measure the Chamfer distance between the predicted and target curve segments as a measure of curve prediction fidelity.
Since the Chamfer distance may be susceptible to degenerate solutions (the curve folding in on itself), we also measure the distance between the endpoints of the target and predicted curves.
Finally, we evaluate the C2C refined errors, 
to assess whether and to what extent curve prediction can refine the point prediction. Our results are summarized in Table.~\ref{tab:ablation}.
%
We observe that larger segment sizes provides more context to the curve prediction task, leading to lower Chamfer errors and lower endpoint errors.
However, a larger segment provides more `landing surface area' (\ie less localization) to a prediction from the P2P model, so the projected point may not lie very close to the target point.
Smaller segments lead to better localization of the P2P points, leading to better C2C-Refined scores.

\begin{table}[t]
    \centering
    \begin{tabular}{c|cc} \hline \hline
        Metric  & \textbf{10 points} & \textbf{20 points} \\ \hline 
         P2P& \multicolumn{2}{c}{9.61} \\  \hline
         C2C-Refined & 7.13 & 7.20 \\
         C2C-Endpoint & 7.65 & 5.83 \\
         C2C (Chamfer) & 4.65 & 3.14 \\
         \hline \hline
    \end{tabular}
    \caption{ \textbf{The effect of segment length.} The numbers shown here are the mean error (mm) of all centerline points.}
    \label{tab:ablation}
\end{table}

\vspace{-3pt}
\subsection{The Shortest Path of Branch Tracing}
\begin{figure}[!t]
\centering
\includegraphics[width=0.45\textwidth]{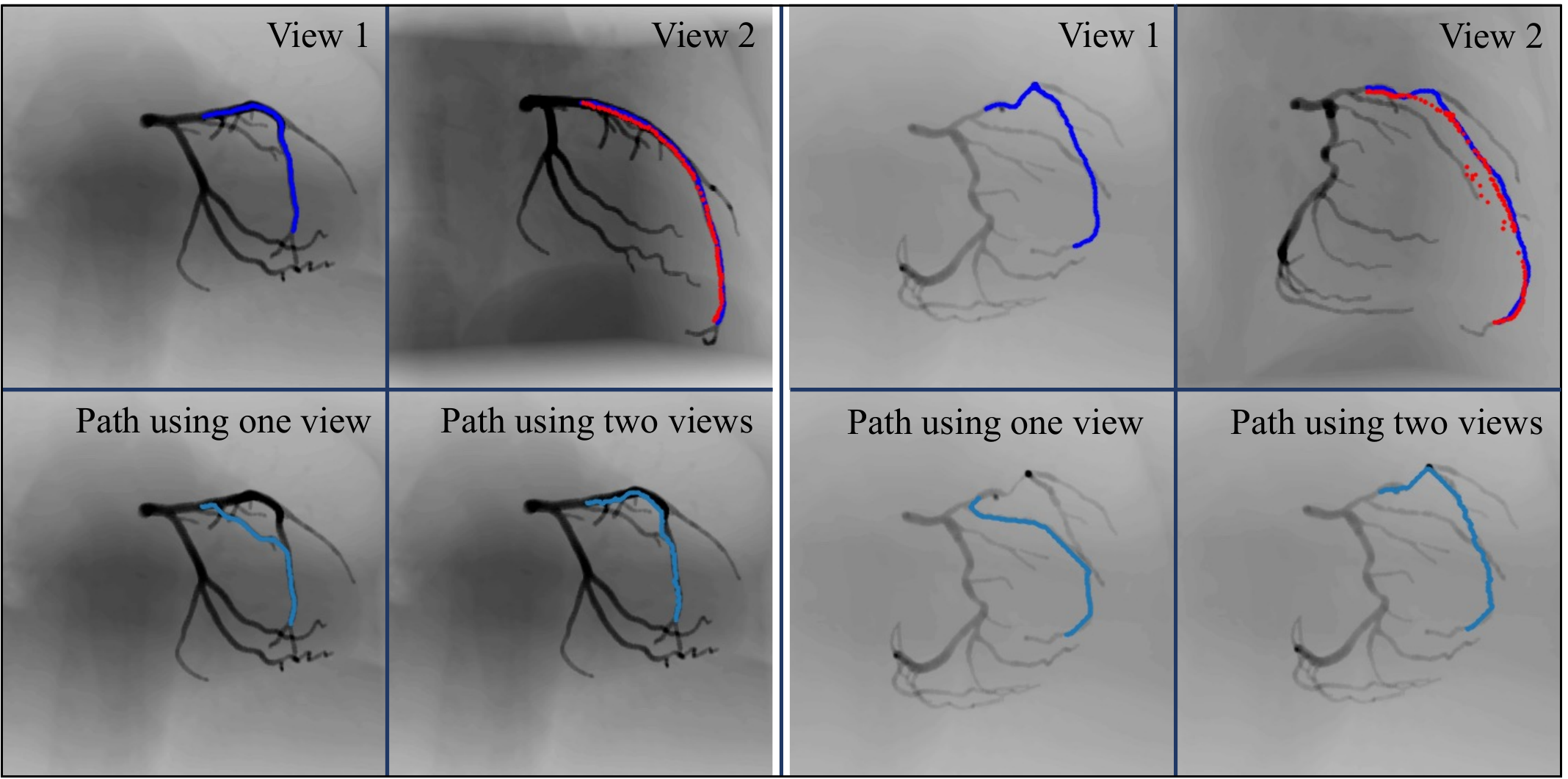}
\caption{\textbf{The shortest path tracing using one vs. two views.} The first row displays view 1 with the ground truth target branch, and view 2 with the predicted corresponding centerline points, where the red color represents the prediction and the blue color represents the ground truth. The second row shows the shortest path given only view 1 in the first column and the shortest path derived by fusing view 2 in the second column. 
}
\label{fig: shortestpath}
\end{figure}

In this section, we analyzed whether centerline tracing can be improved by taking correspondences from the second view into account. For centerline tracing, we computed the shortest path using Dijkstra’s algorithm. First, we computed Frangi’s filter \cite{frangi1998multiscale} on both views to build cost maps. Second, we determined two seed points in the first view on a selected branch. Third, the Dijkstra algorithm was started from the first seed point to compute the shortest path to the second seed point. We conducted two separate experiments: one using the cost only from the first view and the second one using the combined cost from the two views. In the latter case, we predicted the correspondence of the visiting point on the second view using our trained model. As shown in Fig.~\ref{fig: shortestpath}, 
by fusing the additional view with our predictions of correspondence, the experiment produced better traces, especially, avoiding shortcuts around overlapping branches that can falsely appear as bifurcations.

\begin{figure}[!t]
\centering
\includegraphics[width=0.45\textwidth]{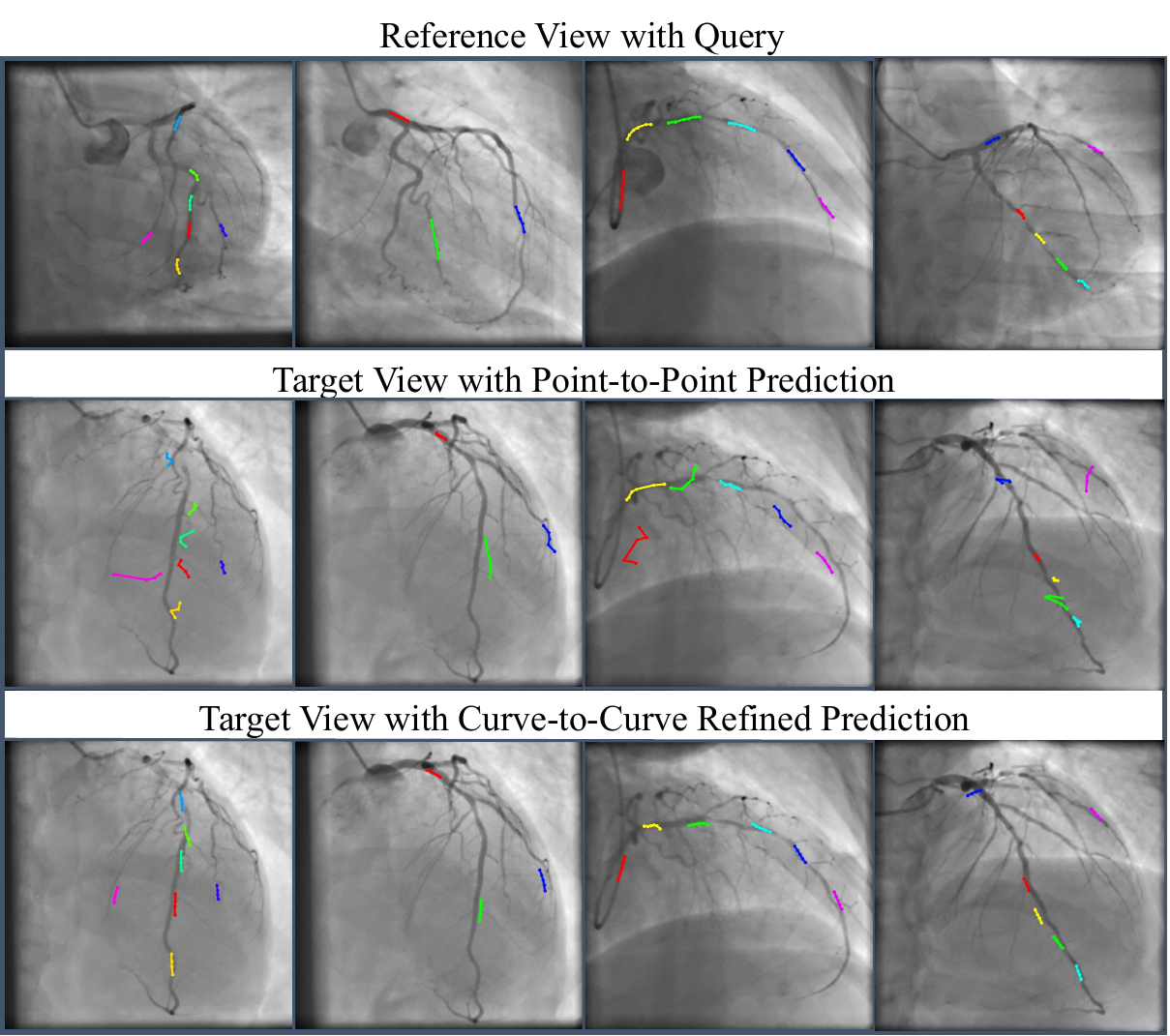}
\caption{\textbf{Qualitative evaluations on the real angiograms.} 
The top row displays the reference view, showcasing the manual queries received through the GUI. The middle row illustrates the target view with predicted correspondences from the Point-to-Point model. The bottom row presents the target view with predicted correspondences, refined by the Curve-to-Curve model.}
\label{fig: real-angio}
\end{figure}

\vspace{-3pt}
\subsection{Qualitative Evaluation on Real Angiograms}
Real angiograms deviate significantly from the simulated angiograms in terms of intensity distribution, background artifacts, catheters, and border effects.
Fig.~\ref{fig: real-angio} shows an initial qualitative evaluation of the trained models on real angiograms, as we lack correspondence annotations.
The source and target views differ a lot in these examples, making the problem challenging. We implemented a GUI that allows users to click a set of points on a query image. These points are then used to predict the corresponding target points or curves, depending on the model used. Figure~\ref{fig: real-angio} demonstrates that the P2P model can predict target correspondences near the correct locations but lacks awareness of the vessel structure in its predictions. This leads to the observation that predictions of sequential query points on a segment do not align with the curvature structures in the images. In contrast, the C2C model consistently projects these points onto vessel structures within the image. Future work aims to perform a quantitative evaluation and narrow the gap between simulated images and real angiograms.

\section{Conclusion and Future Work}
\label{sec:conlcusion}
\vspace{-5pt}
In this study, we introduce a novel problem formulation for establishing dense correspondences between different angiography views. To this end, we developed a point query matching framework. To further reduce ambiguity and improve topological awareness, we extended the framework to curve matching. The effectiveness of our framework has been demonstrated via both quantitative and qualitative analyses, and in downstream tasks such as branch tracing.

For future work, the first area of focus is adapting this approach for use with real angiograms, which requires efficient fine-tuning on real angiograms with sparse labels. The second area involves adding cardiac motion to our data simulation and then incorporating temporal cues in solving correspondence.
{
    \small
    \bibliographystyle{ieeenat_fullname}
    \bibliography{main}
}

\end{document}